\documentclass[final]{cvpr}

\usepackage{times}
\usepackage{epsfig}
\usepackage{graphicx}
\usepackage{amsmath}
\usepackage{amssymb}
\usepackage{amsthm}
\usepackage{subfiles}
\usepackage{booktabs}
\usepackage{arydshln}
\usepackage{tabularx}
\usepackage{colortbl}

\usepackage[utf8]{inputenc}
\usepackage{algpseudocode}
\usepackage{algorithmicx}
\usepackage{algorithm}

\usepackage{graphicx}
\usepackage{adjustbox}
\usepackage{float}
\usepackage{subfigure}

\theoremstyle{definition}
\newtheorem{definition}{Definition}[section]

\newcommand\blfootnote[1]{%
  \begingroup
  \renewcommand\thefootnote{}\footnote{#1}%
  \addtocounter{footnote}{-1}%
  \endgroup
}

\usepackage[pagebackref=true,breaklinks=true,colorlinks,bookmarks=false]{hyperref}

\def\cvprPaperID{5056} 
\def\confYear{CVPR 2021}
\begin{document}

\title{Perceptual Indistinguishability-Net (PI-Net): \\
Facial Image Obfuscation with Manipulable Semantics}

\author{Jia-Wei Chen$^{1,2,\ast}$ \hspace{3mm} Li-Ju Chen$^{2,\ast}$ \hspace{3mm} Chia-Mu Yu$^{3}$ \hspace{3mm} Chun-Shien Lu$^{1,2}$\\
$^{1}$IIS, Academia Sinica \hspace{3mm} $^{2}$CITI, Academia Sinica \hspace{3mm} $^{3}$National Yang Ming Chiao Tung University\\

{\tt\small \{jiawei, lijuchen\}@citi.sinica.edu.tw \hspace{3mm} chiamuyu@nycu.edu.tw \hspace{3mm} lcs@iis.sinica.edu.tw}
}

\maketitle

\begin{abstract}

With the growing use of camera devices, the industry has many image datasets that provide more opportunities for collaboration between the machine learning community and industry. However, the sensitive information in the datasets discourages data owners from releasing these datasets. Despite recent research devoted to removing sensitive information from images, they provide neither meaningful privacy-utility trade-off nor provable privacy guarantees. In this study, with the consideration of the perceptual similarity, we propose perceptual indistinguishability (PI) as a formal privacy notion particularly for images. We also propose PI-Net, a privacy-preserving mechanism that achieves image obfuscation with PI guarantee. Our study shows that PI-Net achieves significantly better privacy utility trade-off through public image data.
\vspace*{-2.0em}
\end{abstract}

\blfootnote{$\ast$ The first two authors contributed equally to this work
\vspace*{-.5em}}
\section{Introduction}

More and more facial image datasets are becoming available in a wide variety of communities. The availability of these datasets presents enormous opportunities for collaboration between data owners and the machine learning community (e.g., social relation recognition \cite{socialrelationrecognition}). However, the inherent privacy risk of such datasets prevents data owners from sharing the data. For example, Microsoft's facial image dataset MS-Celeb-1M, Duke's MTMC, and Stanford's Brainwash were taken down due to the potential privacy concern \cite{NYTIMES, FREEDOMTOTINKER}. 

\textbf{Facial Image Obfuscation.}
Many facial image obfuscation (aka. face anonymization and face de-identification) solutions have been proposed. An approach to mitigating privacy risk of releasing facial images is to use Generative Adversarial Networks (GANs) to synthesize visually similar images \cite{Choi2018StarGANUG, Karras2019ASG}.
Recently, GANs that can manipulate semantics have been proposed to even have a fine-grained control of attributes such as age and gender \cite{He2019AttGANFA, Qian2019MakeAF, Wu2019RelGANMI}. 
Based on inpainting \cite{Li2017GenerativeFC, Sun2018CVPR, Yeh2017SemanticII}, one can also anonymize the image, while retaining the facial semantics, by removing the region of interest (ROI) with sensitive semantics in the image and restoring it.
Image forgery methods such as Deepfakes \cite{Deepfakes}, if used to replace sensitive semantics, can also mitigate privacy risks for identity disclosure \cite{Afchar2018MesoNetAC,Guera2018DeepfakeVD}. 

However, all of the above methods share a common weakness of syntactic anonymity, or say, lack of formal privacy guarantee. Recent studies report that obfuscated faces can be re-identified through machine learning techniques \cite{WIRED, Hill2016PETS, Oh2016ECCV}. Even worse, the above methods are not guaranteed to reach the analytical conclusions consistent with the one derived from original images, after manipulating semantics. To overcome the above two weaknesses, one might resort to differential privacy (DP) \cite{Dwork2014TheAF}, a rigorous privacy notion with utility preservation. In particular, DP-GANs \cite{Abadi2016DeepLW,chen2020gs,Jordon2019PATEGANGS,Torkzadehmahani2019DPCGANDP} shows a promising solution for both the provable privacy and perceptual similarity of synthetic images. Unfortunately, DP-GANs can only be shallow, because of its rapid noise accumulation, hindering model accuracy. One can also apply DP to the image, leading to pixelation \cite{Fan2019DifferentialPF}. 
Consequently, the images generated in such a way are of low quality. 

\textbf{Key Insights.}
Basically, anonymizing facial images, while retaining necessary information for tasks such as detection, recognition, and tracking is very challenging. Notably, our result is in possession of the following novelties. 

First, based on metric privacy \cite{chatzikokolakis2013broadening}, we introduce perceptual indistinguishability (PI), a variant of DP with a particular consideration of perceptual similarity of the facial images. More specifically, PI, while retaining perceptual similarity, achieves the indistinguishability result that an adversary, when seeing an anonymized image, can hardly infer the original image, thereby protecting the privacy of the image content. On the other hand, inherited from DP, PI can also ensure high data utility (detection, classification, tracking, etc.). As far as we know, this is the first time perceptual similarity or more concretely, facial attributes, is used to define an indistinguishability notion from the image adjacency point of view in the context of DP, which enables the reconciliation among privacy and utility. Although facial attributes have been also exploited for face de-identification \cite{li2019anonymousnet,Xiao2020AdversarialLO}, in addition to  relying on different privacy notions, our study is different from theirs in that (1) \cite{li2019anonymousnet} needs pre-processing for face alignment and cropping and (2) \cite{Xiao2020AdversarialLO} learns privacy representation instead of releasing private image dataset.

Second, we introduce PI-Net, a novel encoder-decoder architecture to achieve image obfuscation with PI. PI-Net is featured by its operations in latent space and can also be seen as a latent-coded autoencoder. In particular, we inject noise to the latent code derived from GAN inversion. PI-Net is also featured by the use of triplet loss that clusters the faces with similar facial attributes. This novel architecture enables the network to create anonymized images that look realistic and satisfy user-defined facial attributes. 

In summary, we make the following contributions:
\begin{itemize}
    \item We present a notion for image privacy, perceptual indistinguishability (PI), defining adjacent images by the perceptual similarity between images in latent space.
    \item We propose PI-Net to anonymize faces with the selected semantic attributes manipulation. The architecture of PI-Net generates realistic looking faces with the selected attributes preservation.
\end{itemize}

\section{Related work}


\textbf{Face Anonymization.} 
Face anonymization has been studied extensively. Straightforward methods such as pixelization, blurring, and masking obviously harm the visual quality. Through the adversarial learning that uses the generator as an anonymizer to modify sensitive information and the discriminator as a facial identifier, the trained generator can synthesize high-quality facial images with different identities \cite{Ren2018LearningTA}. On the other hand, many GAN-based face anonymization methods were proposed. For example, the facial attributes of an image are modified via GANs such that the distribution of facial attributes matches the desired $t$-closeness in Anonymousnet \cite{Li2007tClosenessPB,li2019anonymousnet}. One can generate faces via GAN such that the probability of a particular face being re-identified is at most $1/k$ in $k$-same family of algorithms \cite{ Gross2006ModelBasedFD, Jourabloo2015AttributePF, Newton2005PreservingPB}. DeepPrivacy \cite{deepprivacy2019} and CIAGAN \cite{maximov2020ciagan} are conditional GANs (CGANs), generating anonymized images. The former is based on the surrounding of the face and sparse pose information, while the latter relies on an identity control discriminator. Gafni et al.'s proposal \cite{Gafni2019LiveFD} is an adversarial autoencoder, coupled with a trained face-classifier. However, their anonymized images, while successfully fooling the recognition system, are unlikely to hide the identity of the presented faces. 

\textbf{Differentially Private Machine Learning.}
Model inversion attack \cite{modelinversion} aims to reconstruct the training dataset by accessing or querying a model. Defenses against model inversion include DP and adversarial learning \cite{Xiao2020AdversarialLO}. Here, we pay our attention only to DP-GAN \cite{Abadi2016DeepLW,chen2020gs,Jordon2019PATEGANGS,Torkzadehmahani2019DPCGANDP}, as it can generate synthetic images with reasonable data utility. 
Since only the discriminator has access to the training data, one cannot distinguish whether an individual training data participated in training after injecting DP noise into the gradients.
Moreover, variants of DP (e.g., Renyi-DP \cite{Mironov2017RnyiDP}) and different network structures (e.g., \cite{Torkzadehmahani2019DPCGANDP}) can be employed to generate images with higher quality. The $1$-Lipschitz property of the Wasserstein distance is also helpful in determining clipping bounds in gradient sanitization, further improving image quality \cite{chen2020gs}. Unfortunately, DP-GANs use standard DP \cite{Dwork2014TheAF} to define image adjacency and only cares the accuracy of the model fed by synthetic images. Therefore, one can only expect low perceptual similarity between the original and synthetic images, not to speak of the attribute preservation.

\textbf{Differentially Private Multimedia Analytics.}
There is little attention to multimedia analytics with DP. 
Until recently, the first study of defining image adjacency was proposed \cite{Fan2019DifferentialPF}. More specifically, by treating an image as a database, and by treating the features within the image (e.g., pixels) as database records, PixDP \cite{Fan2019DifferentialPF} defines image adjacency with standard DP. DP noise injection in the pixel domain obviously brings catastrophic consequence on image quality, and so the follow-up method \cite{Fan2019PracticalIO} defines image adjacency based on the singular vectors of the SVD transformed images. SVD-Priv \cite{Fan2019PracticalIO} slightly improves the quality of anonymized images. 
In addition to images, DP can also be applied to video in that video adjacency is defined according to whether a sensitive visual element appears in the video \cite{wang2020videodp}. 
Unfortunately, as video adjacency is defined in pixel domain, the video quality will be largely destroyed.

\section{Preliminary}
DP \cite{Dwork2014TheAF} provides a provable guarantee of privacy that can be quantified and analyzed, and the adjacency is a critical concept that defines the information to be hidden. Unfortunately, applying DP, together with Laplace noise \cite{Dwork2014TheAF}, to images leads to unacceptable quality.  
In our work, inspired by metric privacy \cite{chatzikokolakis2013broadening}, we define image adjacency as the difference in the high-level features learned by GANs; \emph{i.e.,} the latent codes, used to capture changes in image semantics.
In other words, we attempt to find the best-suited latent code in the latent space for a given image. 
Then, we give a clear semantic definition of the latent space so that the semantics of the image can be captured for privacy analysis.
Below we present formal definitions of the above terminologies that will be used throughout this paper. The notations frequently used in this paper are described in in Notation Table in the Appendix.

\begin{definition}
    \textbf{(Semantic Space \cite{Shen2020InterpretingTL})} Suppose a semantic scoring function defined as $f_{S}: \mathcal{I} \rightarrow \mathcal{S}$ can evaluate each image's facial attribute components, such as old age, smiling, where $\mathcal{S} \subset \mathbb{R}^{m}$ is called the semantic space formed by $m$ facial attributes (In our case, for a semantic label $s \in \mathcal{S}$, it is a binary vector with each entry indicating whether an attribute exists or not), and $\mathcal{I}$ is the image space. Moreover, given a well-trained GAN model, we denote a generator as $G: \mathcal{X} \rightarrow \mathcal{I}$, such that the latent space $\mathcal{X} \subset \mathbb{R}^{d}$ can be bridged with the semantic space $\mathcal{S}$ through $s = f_{S}(G(x))$, and so the semantic of the latent code $x$ can be evaluated.
    \label{def:semantic}
\end{definition}

\begin{definition}
    \textbf{(GAN Inversion \cite{Abdal2019Image2StyleGANHT})} Given a trained GAN model, the goal of GAN inversion is to find the most accurate latent code $x^{\prime} \in \mathcal{X}$ to recover the input image $I \in \mathcal{I}$. In our case, an optimization-based method is used to minimize the reconstruction error of the features between the generated image and the given image by directly optimizing the latent code $x^{\prime}$. We use $F$ to denote the optimization of this inversion: 
    $F: \mathcal{I} \rightarrow \min _{\mathbf{x^{\prime}}} \mathcal{L} =\left\| \mathcal{I} - G\left(\mathbf{x}^{\prime}\right)\right\|_{2} + \alpha \cdot \|
   V(\mathcal{I}) - V(G\left(\mathbf{x}^{\prime}\right))\|_{2}$ where $V$ is the VGG16 \cite{Simonyan2015VeryDC}.
    \label{def:inversion}
\end{definition}

\begin{definition}
    \textbf{($d_{\mathcal{X}}$-Privacy \cite{chatzikokolakis2013broadening})} A randomized mechanism $K: \mathcal{X} \rightarrow \mathcal{P}(\mathcal{Z})$ satisfies $d_{\mathcal{X}}$-privacy, iff $\forall x, x^{\prime} \in \mathcal{X}$: 
    \begin{equation}
        K(x)(Z) \leq e^{d_{\mathcal{X}}\left(x, x^{\prime}\right)} K\left(x^{\prime}\right)(Z), \quad \forall Z \in \mathcal{F}_{\mathcal{Z}},
    \end{equation}
    where $d_{\mathcal{X}}$ is a distance metric for $\mathcal{X}$, $\mathcal{Z} \subseteq \mathcal{X}$ is a set of query outcomes over $\mathcal{X}$, $\mathcal{F}_{\mathcal{Z}}$ is a  $\sigma$-algebra over $\mathcal{Z}$, and $\mathcal{P}(\mathcal{Z})$  is the set of probability measures over $\mathcal{Z}$.
    \label{def:dx}
\end{definition}

Here, metric privacy \cite{chatzikokolakis2013broadening} generalizes DP to protect secrets in an arbitrary domain $\mathcal{X}$. Essentially, based on a distance metric $d_{\mathcal{X}}$, $K$ ensures a level of indistinguishability proportional to the distance. An adversary observing the outputs (e.g., $Z$) can hardly infer the exact input.

\begin{definition}
\textbf{($\Delta$-Sensitivity \cite{chatzikokolakis2013broadening})}
A deterministic function $f : \mathcal{X} \rightarrow \mathcal{Y}$ is $\Delta$-sensitive with respect to $d_{x}$ and $d_{y}$ iff $d_{y}\left(f(x), f\left(x^{\prime}\right)\right) \leq \Delta d_{x}\left(x, x^{\prime}\right)$ for all $x, x^{\prime} \in \mathcal{X}$. 
The smallest $\Delta$ is called the sensitivity of $f$ with respect to $d_{x}$ and $d_{y}$.
\label{def:sensitive}
\end{definition}

\begin{definition}
    \textbf{($\Delta d_{\mathcal{X}}$-Privacy \cite{chatzikokolakis2013broadening})}
    Assume that $f$ is $\Delta$-sensitive with respect to $d_{x}$ and $d_{y}$, and the mechanism $H$ satisfies $d_{\mathcal{Y}}$-privacy. Then, the mechanism $M: \mathcal{X} \rightarrow \mathcal{P(Z)}$ defined as $M(x) = (H \circ f)(x)$ satisfies $\Delta d_{\mathcal{X}}$-privacy, where $\circ$ is the composition operator.
    \label{def:ddx}
\end{definition}

\section{Our Solution: PI-Net}

In this section, we detail our image obfuscation mechanism based on perceptual distance.
First, we formalize our proposed privacy notion, perceptual indistinguishability (PI).
Then, we describe our proposed semantic transformation for manipulating the chosen facial attributes.
Finally, we illustrate the mechanism used to obfuscate the identifiable information on the face.

\subsection{Privacy Formulation}
Standard DP \cite{Dwork2014TheAF} has been a de facto standard for data privacy, and widely used to protect the privacy of tabular (non-multimedia) data. Nevertheless, tabular data (composed of records) intrinsically are different from multimedia data (composed of low-level features like pixels or composed of high-level features like semantic features).
Thus, standard DP is not applicable to privacy protection of multimedia, and a privacy notion for multimedia is desirable.

In an image, there are many facial attributes that can be captured by latent codes learned from GANs.
For these latent codes that can correctly represent attributes, an interpretation is that they correctly learn the semantic information embedded in the image and have the corresponding semantic scores. 
We are particularly interested in semantic information that can also be used to represent an individual's identity.
Studies in semantic editing \cite{Abdal2019Image2StyleGANHT,Shen2020InterpretingTL} have concluded that the semantics captured in the latent code are related to individual identity and can be used to change the identity of the individual visually.
Therefore, one can expect a properly-designed distance function that can measure the disimilarity between latent codes, and use it to protect an individual identity.

\begin{definition}
\textbf{(Perceptual Distance)}
Given a well-trained GAN model, the Euclidean distance $\rho(x, x^{\prime})$ between two latent codes $x$ and $x^{\prime}$ is called the perceptual distance, where $x, x^{\prime} \in \mathcal{X}$, and $\mathcal{X}$ is latent space learned in GANs.
\label{def:p-distance}
\end{definition}
Perceptual distance is a metric for measuring perceptual similarity between images. Studies \cite{Gafni2019LiveFD,johnson2016perceptual,Xiao2020AdversarialLO} have proposed a variety of definitions for perceptual distance, with an attempt to improve image quality or privacy protection. Despite no standardized definition for perceptual distance, they all can be seen as a distance between high-level features learned from neural networks.
In our work, we adopt Euclidean distance between latent codes learned from GANs model as the perceptual distance. There are two reasons behind such a design choice. The first comes with the constraint that the distance in metric privacy is required to satisfy the triangular inequality.
Second, as the latent space of GAN models has arithmetic properties \cite{Shen2020InterpretingTL}, our use of Euclidean distances, together with a proper noise injection, suffices to alter individual identities.

In the following, we use the $d_\mathcal{X}$-privacy in Definition \ref{def:dx} to formulate a notion of privacy based on perceptual distance. In our consideration, privacy is proportional to semantic information; \emph{i.e.,} images that are semantically similar are more indistinguishable from each other, and thus have a higher probability of generating the same obfuscated output. Following the above principle, we define the image adjacency and privacy notion below.

\begin{definition}
\textbf{(Adjacent images)} 
Two facial images are adjacent, if the perceptual distance between the corresponding latent codes $\rho(x, x^{\prime}) \leq \beta$. As $\beta$ goes smaller, two faces will have more similar facial attributes, while the identities will also be similar.
\end{definition}

\begin{definition}\label{def:pi}
    \textbf{($\epsilon$-Perceptual Indistinguishability)} 
    A randomized mechanism $K$ satisfies $\epsilon$-Perceptual Indistinguishability ($\epsilon$-PI) if $\forall x, x^{\prime} \in \mathcal{X}$ that are adjacent :
    \begin{equation}
        \ln \left| \frac{K(x)(Z)}{K(x^{\prime})(Z)} \right| \leq \epsilon \rho(x, x^{\prime} ),
    \end{equation} where $K(x)(Z)$ is the probability that the obfuscated latent code belongs to the set $Z \subseteq \mathcal{X}$ when the original latent code is $x$, and $K(x)$ is the probability distribution over $\mathcal{X}$.
\end{definition}
An interpretation of Definition \ref{def:pi} is that the closer the perceptual distance between the latent codes (the more similar the semantic information measured), the closer the probability of producing the same obfuscated output, thus making it more difficult for an adversary to distinguish between true codes.

\begin{figure*}[h]
    \centering
    \includegraphics[width=0.8\linewidth]{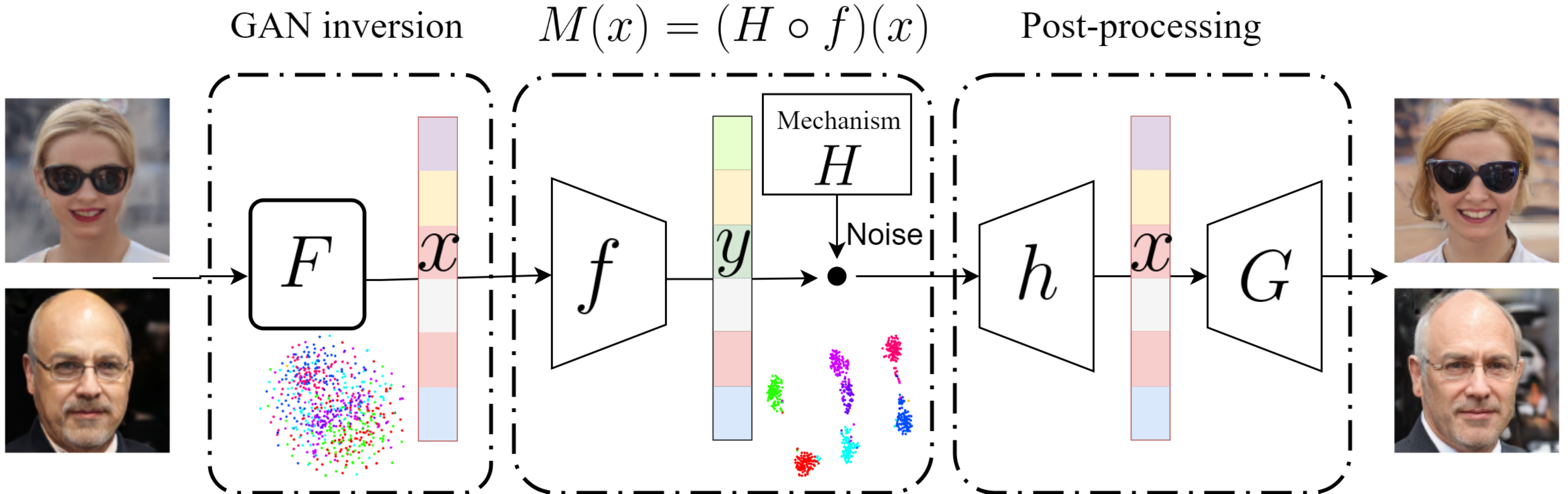}
    \vspace{2mm}
    \caption{PI-Net: (Left dash-dot block) GAN inversion, (Middle dash-dot block) Encoding network, and (Right dash-dot block) Decoding network and GAN. The mechanism $H$ is not required during training. The $\mathcal{X}$ space and $\mathcal{Y}$ space are visualized with $t$-SNE \cite{Maaten2008VisualizingDU}.}
    \label{fig:arch}
\end{figure*}

\subsection{Training PI-Net}
An ideal facial image obfuscation would obfuscate the latent code's identifiable information without interfering with other facial attributes.
Nevertheless, studies \cite{Abdal2019Image2StyleGANHT,Shen2020InterpretingTL} have observed that facial attributes are entangled in the latent codes learned from GANs. Consequently, adding noise directly to the latent codes will also interfere with the facial attributes we desire to preserve, resulting in the poor utility of the obfuscated images.
For example, the two attributes of ``old age'' and ``wearing glasses'' are entangled in the latent code. When we desire to preserve the ``wearing glasses'' attribute, but once the noise is added to the ``old age'', the ``wearing glasses'' attribute is changed by the unwanted interference.

\textbf{Transform.} We propose a transformation framework that can cluster the attribute information we desire to preserve from the entangled latent space, allowing the mechanism to determine the strength of the obfuscation without interference with other attributes for better data utility.

The overall architecture is shown in Figure \ref{fig:arch}, and can be separated into three components: (1) GAN inversion, (2) Encoding network, and (3) Decoding network and GAN.
The operation path is as follows: 
First, the image $I \in \mathcal{I}$ is inverted to an appropriate code $x$ in the latent space $\mathcal{X}$ by Definition $\ref{def:inversion}$. 
The next step is to cluster the codes with the same specified facial attributes through the encoding network $f$, which serves as a transformation.
In the last step, the transformed code in the $\mathcal{Y}$ space is mapped back to a latent space $\mathcal{X}$ using the decoding network $h$, and can then be handed over to the generator $G$ for image synthesis.
Note that mechanism $H$ presented in Figure \ref{fig:arch} is not required in the training procedure and will be described in Section \ref{sec:Mechanism (Prediction Procedure)}. 

The key idea is that, given a semantic defined in Definition \ref{def:semantic} and the specified attributes to be preserved, the transformed codes $y=f(x)$ with the same specified attributes should be close to each other, and the other transformed codes should be far from each other, as illustrated in Figure \ref{fig:arch}.
As the transformed codes have been clustered by the specified attributes, each transformed code is surrounded by other codes with the same specified attributes. As a consequence, adding noise from the distribution $(H \circ f)(x)$ results in a code that preserves the same attributes but changes its identity.

\textbf{Triplet Training.}
When training the encoding network $f$, we use the triplet loss shown in Equation \ref{eqn:triplet} as:
\begin{equation}
\begin{split}
L_{triplet}(x, \theta) = \sum_{i=1}^{N} ||f_{\theta}(x_{i}^a)-f_{\theta}(x_{i}^p)||^2_2 \\ -||f_{\theta}(x_{i}^a)-f_{\theta}(x_{i}^n)||^2_2 + \mu.\\
\end{split}
\label{eqn:triplet}
\end{equation}

In the triplet loss training, given a semantic formed by $m$ attributes, a triplet $(x_{i}^{a}, x_{i}^{p}, x_{i}^{n})$ contains two latent codes of the same specified attributes, called anchor $x_{i}^{a}$ and positive sample $x_{i}^{p}$, and a third latent code must be from different attributes, called negative sample $x_{i}^{n}$. 
The triplet loss function requires the anchor point $x_{i}^{a}$ to be closer to the positive sample $x_{i}^{p}$ than the negative sample $x_{i}^{n}$, and $\mu$ is a threshold that encourages the negative sample to be further away from the anchor point than the positive sample.

Then, we need to map the code in the space $\mathcal{Y}$ back to the latent space $\mathcal{X}$ as close as possible, allowing the generator to synthesize similar images.
Thus, we train a decoding network $h_{\omega}$ by using Equation \ref{eqn:mse} as: 
\begin{equation}  
L_{recon}(x, \omega) = \sum_{i=1}^{N}||x_i-h_{\omega}(f_{\theta}(x_i))||_2,
\label{eqn:mse}
\end{equation} which requires that the transformed code $h_{\omega}(f_{\theta}(x_i))$ must be similar to the original code $x$. 

Finally, since triplet loss only considers the clustering of codes with the same specified attributes, irrelevant to classification results. Therefore, we additionally penalize misclassification by the cross entropy loss that is defined as:
\begin{equation}  
L_{ce}(x, \theta, \omega) = -\sum_{i=1}^{N}s_{i} \log \left(f_A\left(h_{\omega}\left(f_{\theta}\left(x_{i}\right)\right)\right)\right),
\label{eqn:ce}
\end{equation}
where $f_A$ is the trained attribute classification
model, and $s_{i} \in \mathbb{R}^{m}$ is the corresponding correct semantic, which is a vector formed by $m$ binary attributes. 

\begin{algorithm}[t]
\caption{Training PI-Net} 
\hspace*{0.02in} {\bf Input:} \\    
\hspace*{0.15in} Images: $\mathcal{I} = \{I_1,...,I_N\}$\\
\hspace*{0.15in} Semantic space: $\mathcal{S} = \{s_1,...,s_N\}$, where $s_i \in \mathbb{R}^{m}$\\
\hspace*{0.15in} Learning rate: $\alpha, \beta$.\\
\hspace*{0.02in} {\bf Output:} \\    
\hspace*{0.15in} Model weights: $\theta$, $\omega$
\vspace{0,04in}

\begin{algorithmic}[1]
\State $\mathcal{D} =\{(I_1,s_1),...,(I_N,s_N)\}$
\For{$I_i$ in $\mathcal{I}$}
    \State Invert the image $I_i$ to get the latent code $x_i$ \par 
    using $x_i = F(I_i)$ and $F$ in Definition $\ref{def:inversion}$
\EndFor
\While {not done} 
    \State Shuffle $\{(x_1, s_1), \ldots , (x_N, s_N)\} \subseteq \mathcal{D}$ \vspace{0.01in}
    \State Split the shuffled dataset $D$ to get batches $B_{1, \dots,m}$ \vspace{0.02in}
    \For{$B_{i}$ in $B_{1 \ldots m}$}
        \vspace{0.01in}
        \State $b_i = |B_{i}|$
        \State {Get $b_i$ datapoints $(x_1, s_1), ... ,(x_{b_i}, s_{b_i}) \in B_{i}$} \vspace{0.01in}
        \State Evaluate $\nabla_{\theta} L_{triplet}$ and $\nabla_{\theta} L_{ce}$ using $B_{i}$  \par \hspace{0.3in} and $L_{triplet}, L_{ce}$ in Equations $\ref{eqn:triplet}$ and $\ref{eqn:ce}$  
        \vspace{0.01in}
        \State Update $\theta \leftarrow \theta - \alpha (\nabla_{\theta} L_{triplet} + \nabla_{\theta} L_{ce})$ \vspace{0.01in}
        
        \State Evaluate $\nabla_{\omega} L_{recon}$ using $B_{i}$ \par \hspace{0.3in} and $L_{recon}$ in Equation $\ref{eqn:mse}$
        \vspace{0.01in}
        \State Update $\omega \leftarrow \omega - \beta \nabla_{\omega} L_{recon}$
        \vspace{0.01in}
    \EndFor
\EndWhile
\end{algorithmic}
\label{alg:overall}
\end{algorithm}

The above neural network training procedures are summarized in Algorithm \ref{alg:overall}.

\subsection{Image Obfuscation via PI-Net}\label{sec:Mechanism (Prediction Procedure)}
As training PI-Net does not involve noise injection mechanism $H$, this section clarifies three issues during image obfuscation:
(1) How to privatize $f$? (2) How to guarantee PI? and (3) The design of noise injection mechanism $H$.

\textbf{How to Privatize $f$.}
As shown in Figure \ref{fig:arch}, when the data owner wants to share the images by a trained PI-Net, the latent codes $x$ are transformed through the encoding network $f: \mathcal{X} \rightarrow \mathcal{Y}$ for high data utility.
To obtain a transformed code from $f$, one can define $M(x) = (H \circ f)(x)$, where $M: \mathcal{X} \rightarrow \mathcal{P(Y)}$ and $H: \mathcal{Y} \rightarrow \mathcal{P(Y)}$. Here, the noise injection mechanism $H$ will be described later in Equation \ref{eqn:sphere}, which assigns a noise sampling from a particular distribution to the transformed code $y$, so as to satisfy $d_\mathcal{Y}$-privacy (see more details later). 
The $d_\mathcal{Y}$-privacy, whose interpretation of $d_\mathcal{Y}$-privacy is the same as Definition \ref{def:dx} but on a different metric space, ensures that the level of privacy provided by $H$ is proportional to the metric $d_{\mathcal{Y}}$ on the $\mathcal{Y}$ space. Note that in the framework of metric privacy, we can express the privacy of the ``mechanism'' itself in its own space, without the need to consider the deterministic function or $\Delta$-sensitivity. 

\textbf{How to Guarantee PI.}
When $H$ satisfies $d_\mathcal{Y}$-privacy, $M$ can be proven to achieve $\Delta d_\mathcal{X}$-privacy through $\Delta$-sensitivity, where $\mathcal{X}$ denotes the space of defining the perceptual distance, according to Definition \ref{def:ddx}. The calculation of $\Delta$-sensitivity is shown below.
Based on Definition \ref{def:ddx}, $\Delta$-sensitivity, and $d_\mathcal{Y}$-privacy, we can prove that the mechanism $M$ satisfies $\epsilon$-PI in Definition \ref{def:pi}. The detailed proof is omitted here (Please see Appendix.). 

In particular, the $\Delta$-sensitivity in Definition \ref{def:sensitive} is also known as the Lipschitz continuity. 
Although the neural network can add constraints such as spectral normalization or gradient penalty in the training phase to achieve Lipschitz continuity \cite{Gulrajani2017ImprovedTO, miyato2018spectral}, these methods are not applicable to our case, where each cluster requires its own $\Delta$-sensitivity.
More specifically, as the encoding network $f$ will generate clusters according to the number of specified attributes, we need to calculate each cluster's $\Delta$-sensitivity separately.
Assume there are $J$ clusters in $\mathcal{Y}$ space. In order to bound the sensitivity of encoding network $f$ on code $x_j$ that belongs to $j$-th cluster, $j=1,..,J$, we clip the vector $\boldsymbol{g_j}$ by $\boldsymbol{g_j}/ {\max} \left(1, \|\boldsymbol{g_j}\|/C_{j} \right)$ to ensure $\|\boldsymbol{g_j}\|_2 \leq C_j$,
where $\boldsymbol{g_j} = f(x_{j}) - f(x_{j}^{\prime}) \in \mathbb{R}^{k}$, and $C_{j} = \Delta \|x_{j} - x_{j}^{\prime}\|_2 \in \mathbb{R}$ is the perceptual distance.
This clipping step is also used in DP-GANs \cite{Abadi2016DeepLW,chen2020gs,Torkzadehmahani2019DPCGANDP}. 

\textbf{Generating Noise in $H$.} 
Our mechanism $H$ in Figure \ref{fig:arch} follows the method \cite{Fan2019PracticalIO} proposed for sampling on the hypersphere.
In a $k$-dimensional space $\mathcal{Y}$, the mechanism $H = D_{\epsilon, k}$ samples $y$ given $y_0$ according to the probability density function defined as: 
\begin{equation}
    D_{\epsilon, k}\left(\boldsymbol{y}_{\mathbf{0}}\right)(\boldsymbol{y})=C_{\epsilon, k} e^{-\epsilon \cdot d_{\mathcal{Y}}\left(\boldsymbol{y}_{0}, \boldsymbol{y}\right)},
    \label{eqn:sphere}
\end{equation}
where $d_{\mathcal{Y}}$ is the $k$-dimensional Euclidean distance, and 
\begin{equation}
    C_{\epsilon, k}=\frac{1}{2}\left(\frac{\epsilon}{\sqrt{\pi}}\right)^{k} \frac{\left(\frac{k}{2}-1\right) !}{(k-1) !}.
    \label{eqn:C}
\end{equation}
Equation \ref{eqn:sphere} is a variant of the multivariate Laplace mechanism that satisfies $d_{\mathcal{Y}}$-privacy, which means that the level of indistinguishability provided by the mechanism $H$ is proportional to the $k$-dimensional Euclidean distance. The same way to measure privacy using metrics can also be seen in the Definition \ref{def:dx}, and the mechanism $H \circ f$ satisfies $\epsilon$-PI in Definition \ref{def:pi}, as we prove in the Appendix.

Sampling according to Equation \ref{eqn:sphere} can be achieved as follows.
We first convert the Cartesian coordinates of $y$ to the
hyper-spherical coordinate system with $y_0$ at the origin, resulting in $1$ radial coordinate and $k-1$ angular coordinates. 
Then, combined with the formula for $n$-sphere surface area, we can deduce that the marginal probability of radial is the gamma-beta distribution $f(x)=\frac{\beta^{\alpha}}{\Gamma(\alpha)} x^{\alpha-1} e^{-\beta x}$, where the shape $\alpha$ corresponds to dimension $k$, the ratio $\beta$ corresponds to $\epsilon$, and the variable $x$ corresponds to the radial $r$.
Two steps are taken next:
(1) Sampling radial coordinates according to marginal distributions, \emph{i.e.,} gamma-beta distribution.
(2) Uniformly sample a point on the unit $(k-1)$-sphere.
Since the radial and angle are independent of each other, multiplying the results of (1) and (2) gives the sampled noise.

\section{Experiments}
In this section, we first describe the experimental settings.
After that, we visually compare the obfuscated images between PI-Net and other DP-based methods.
At the end, we perform the evaluation of PI-Net in terms of image quality, attribute re-classification accuracy, and trade-off between privacy and utility.

\subsection{Experimental Setting}
Here, we describe the proposed network model of our method and the datasets used in the experiments.

\textbf{Network Architecture.}
Our proposed PI-Net is composed of three components, including (1) GAN inversion, (2) autoencoder, and (3) generator in trained GANs, as illustrated in Figure \ref{fig:arch}. 
The GAN inversion has shown in Definition \ref{def:inversion}, and the autoencoder is composed of the fully connected layers and the ReLU activation function.
As PI-Net is independent of specific GAN models, we have freedom to employ arbitrary GANs as the generator. 
In our experiment, we used the generator of StyleGAN \cite{Karras2019ASG} in PI-Net, and set StyleGAN's $W$ space as PI-Net's latent space $\mathcal{X}$.
The reason behind such a design choice is that StyleGAN has been studied to be effective in capturing high-level features such as age, smiling and other facial attributes \cite{Abdal2019Image2StyleGANHT,Shen2020InterpretingTL}.
Moreover, the dataset used for training our PI-Net is also the synthetic images that generated by StyleGAN. 
The following describes the dataset used in our evaluations.

\textbf{Training Dataset.} 
Based on a pre-trained StyleGAN model, we randomly sampled and synthesized 50K images from the latent space. 
There are two reasons of adopting this way to generate synthetic dataset: (1) To collect sufficient samples of specific facial attributes to avoid bias from the unbalanced dataset, and (2) To avoid the use of GAN inversion for latent code calculation of real images, saving a considerable amount of time in building the dataset.
In our training dataset generation, since the randomly selected latent codes do not have face attribute labels, we used the CelebA \cite{Liu2015celeba} to train the attribute prediction model based on ResNet-50 \cite{He2019resnet} to assign attribute labels to each sampled latent code.
Meanwhile, to avoid assigning false face labels, we added only the selected samples with sufficiently high confidence levels to the training dataset.

\textbf{Testing Dataset.} 
We used CelebA to evaluate the performance of our trained PI-Net. 
CelebA has facial attribute data from more than 200K celebrity images, with different celebrity identities labeled in the dataset, and each celebrity image has 40 binary attributes.
It has been widely used for different computer vision tasks, including face detection, face recognition, and attribute prediction.
In our experiments, we used the celebrity identities provided by the dataset and the corresponding binary attributes to evaluate our privacy protection and data utility.

\begin{figure}[t]
    \centering
    \includegraphics[width=0.9\linewidth]{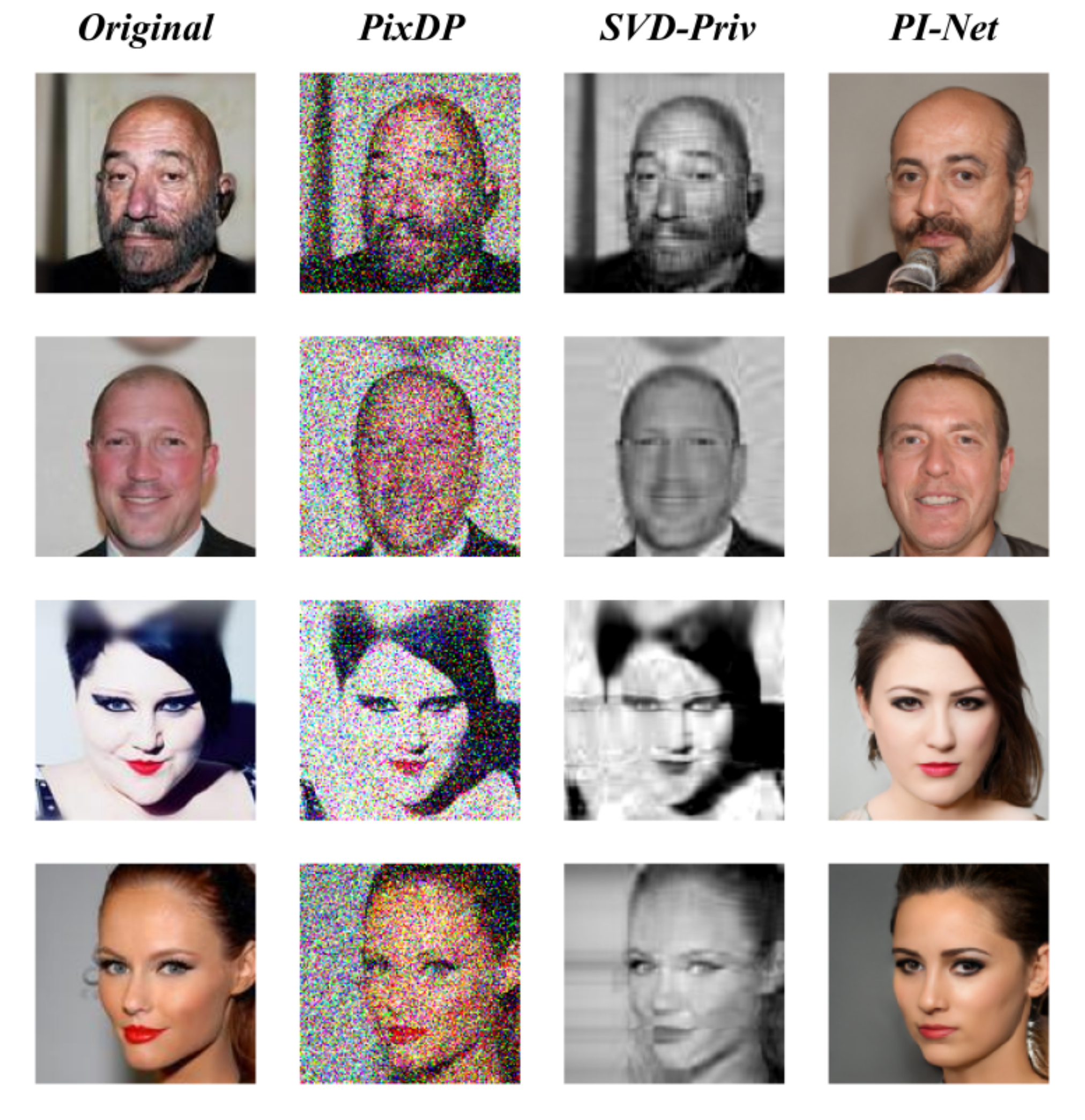}
    \caption{Example images and the corresponding obfuscated results. From left to right, the first column represents original images. PixDP \cite{Fan2019DifferentialPF} (2nd column), SVD-Priv \cite{Fan2019PracticalIO} (3rd column), and PI-Net (4th column) obfuscate low-level feature (pixels), mid-level feature (geometric structures), and high-level feature (image semantics), respectively. To better observe the shape of faces, we actually used less noises on PixDP and SVD-Priv than PI-Net.}
    \label{fig:vis}
\end{figure}

\subsection{Visualization}
We show the visualization results obtained from PI-Net and compare our results with the ones derived from PixDP \cite{Fan2019DifferentialPF} and SVD-Priv \cite{Fan2019PracticalIO}. 
We particularly note that only PixDP and SVD-Priv were chosen for comparison because they are the solutions that provide formal privacy. As can seen from Figure \ref{fig:vis}, PI-Net generates obfuscated images that have much better realistic-looking.

The visual difference among PI-Net, PixDP, and SVD-Priv can be attributed to three factors. 
(1) PixDP defines the indistinguishability of identifiable features on the pixel space in a straighforward manner. Consequently, after adding DP noise, the image will be mostly pixelated, resulting in pessimistic image quality.
(2) SVD-Priv defines the indistinguishability on the singular value space. Since the singular value of an image determines its geometric structure, obfuscating the singular value will interfere with the overall geometry structure of the image.
(3) PI-Net defines the indistinguishability on the high-level feature space, learned from neural networks. 
As high-level features can capture abstract features such as image semantics, compared to the mid-level and lower-level features such as geometric structures and pixels, image quality will not be dramatically degraded after obfuscation.

\subsection{Evaluation}
In addition to its formal privacy guarantee from PI, PI-Net was also evaluated empirically in terms of (1) quality of the obfuscated image, (2) accuracy of attribute re-classification, and (3) trade-off between privacy and utility. 
First, quality of the obfuscated image is used to measure the similarity degradation between the resultant privacy-protected image and its original one in terms of Frechet Inception Distance (FID) \cite{Heusel2017GANsTB} and Structural Similarity Image Measurement (SSIM) \cite{Wang2004ImageQA}.
Second, accuracy of attribute re-classification is used to verify if the attributes, intended to be preserved, can be retained well. 
We used a pre-trained attribute prediction model based on ResNet-50 \cite{He2019resnet} to extract the attributes and calculate the attribute preservation ratio, which is defined as the number of privacy-protected images whose attributes are re-classified correctly divided by the number of total outputs.
Third, we examined the trade-off between privacy and utility of PI-Net, where privacy is measured in terms of face recognition via pre-trained FaceNet \cite{schroff2015facenet} based on Inception-Resnet backbone
\cite{Szegedy2017Inceptionv4IA} in that low recognition rate implies high privacy, and utility is measured in terms of face detection.

\textbf{Quality of The Obfuscated Image.} We can see from Table  \ref{tab:img-quality-tbl} that PI-Net yields the lowest FID and highest SSIM when compared with PixDP \cite{Fan2019DifferentialPF} and SVD-Priv \cite{Fan2019PracticalIO}. The above FID and SSIM results are also consistent with Figure \ref{fig:vis}, where PI-Net generates visually pleasing images.

\begin{table}[h]
\begin{center}
\begin{adjustbox}{max width=0.45\textwidth}
\begin{tabular}{ccccccc}
        && \textbf{SSIM} &&& $\mathrm{\textbf{FID}}$ & \\
\toprule
$\epsilon$        & $\mathrm{\textbf{0.1}}$ & $\mathrm{\textbf{0.5}}$ & $\mathrm{\textbf{1.0}}$ & $\mathrm{\textbf{0.1}}$ & $\mathrm{\textbf{0.5}}$ & $\mathrm{\textbf{1.0}}$ \\
\toprule
PixDP    & 0.01 &   0.02    & 0.03 & 445 & 445 &  442 \\
\midrule
SVD-Priv  & 0.07 & 0.21 & 0.31 & 333 & 271 & 250 \\
\midrule
PI-Net   &0.29 & 0.35 & 0.37 & 61 & 59 & 59 \\
\bottomrule
\vspace{2mm}
\end{tabular}
\end{adjustbox}
\caption{Image quality (higher SSIM and lower FID implies better image quality) vs. privacy budget (lower $\epsilon$ implies better privacy).}
\label{tab:img-quality-tbl}
\end{center}
\end{table}



\begin{figure}[!h]
    \centering
    \includegraphics[width=0.8\linewidth]{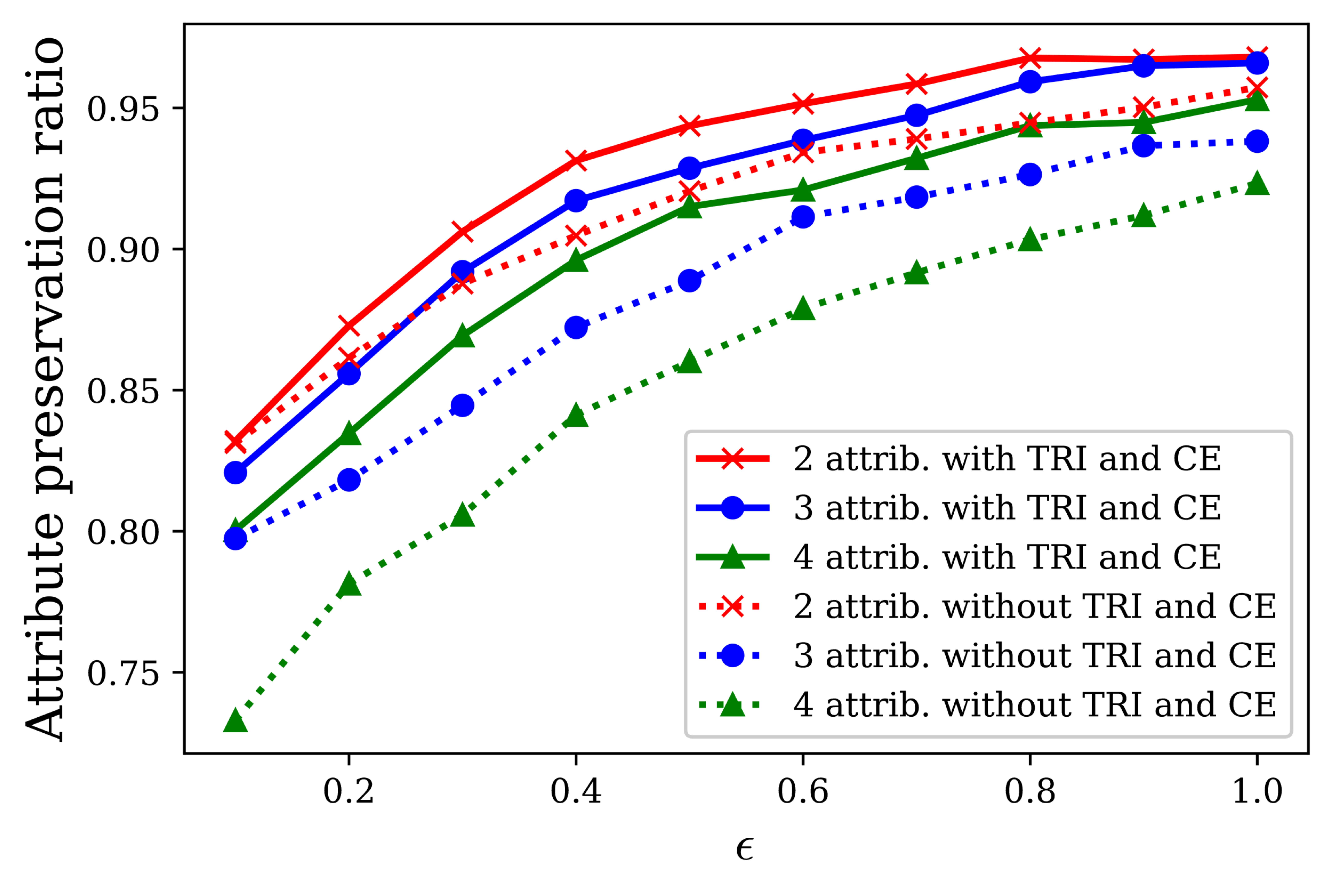}
    \caption{Attribute preservation ratio vs. privacy budget $\epsilon$ with/without triplet (TRI) and cross-entropy (CE) losses.}
    \label{fig:acc}
\end{figure}

\textbf{Accuracy of Attribute Re-Classification.} Figure \ref{fig:acc} shows the attribute preservation ratios under different privacy budgets $\epsilon$ and different numbers of attributes to be preserved.
The face attributes that are specified to be preserved are depicted in Table \ref{tab:attri-tbl}.
The red, blue, and green curves in Figure \ref{fig:acc} denote the results for retaining two-attribute, three-attribute, and four-attributes, respectively. 
The solid curves were generated using the triplet and cross-entropy loss functions but the dash curves were not.
Such an ablation study shows that the use of triplet and cross-entropy loss indeed helps in retaining the desired facial attributes.

\begin{table}[h]
\begin{center}
\scalebox{0.1}{}{
\begin{tabular}{cc}
                      & \textbf{Preserved face semantic}    \\
\toprule                      
2 attributes          & Young, Smiling  \\
\midrule
3 attributes          & Young, Smiling, HeavyMakeup \\
\midrule
4 attributes          & Young, Smiling,  HeavyMakeup \\ 
                      & HighCheekbones \\
\bottomrule
\end{tabular}
}
\vspace{2mm}
\caption{\label{tab:attri-tbl}
Three cases of attributes selected to be preserved.}
\end{center}
\end{table}\vspace{-1mm}

\begin{figure}[h]
    \centering
    \subfigure[]{
        \includegraphics[width=0.3\linewidth]{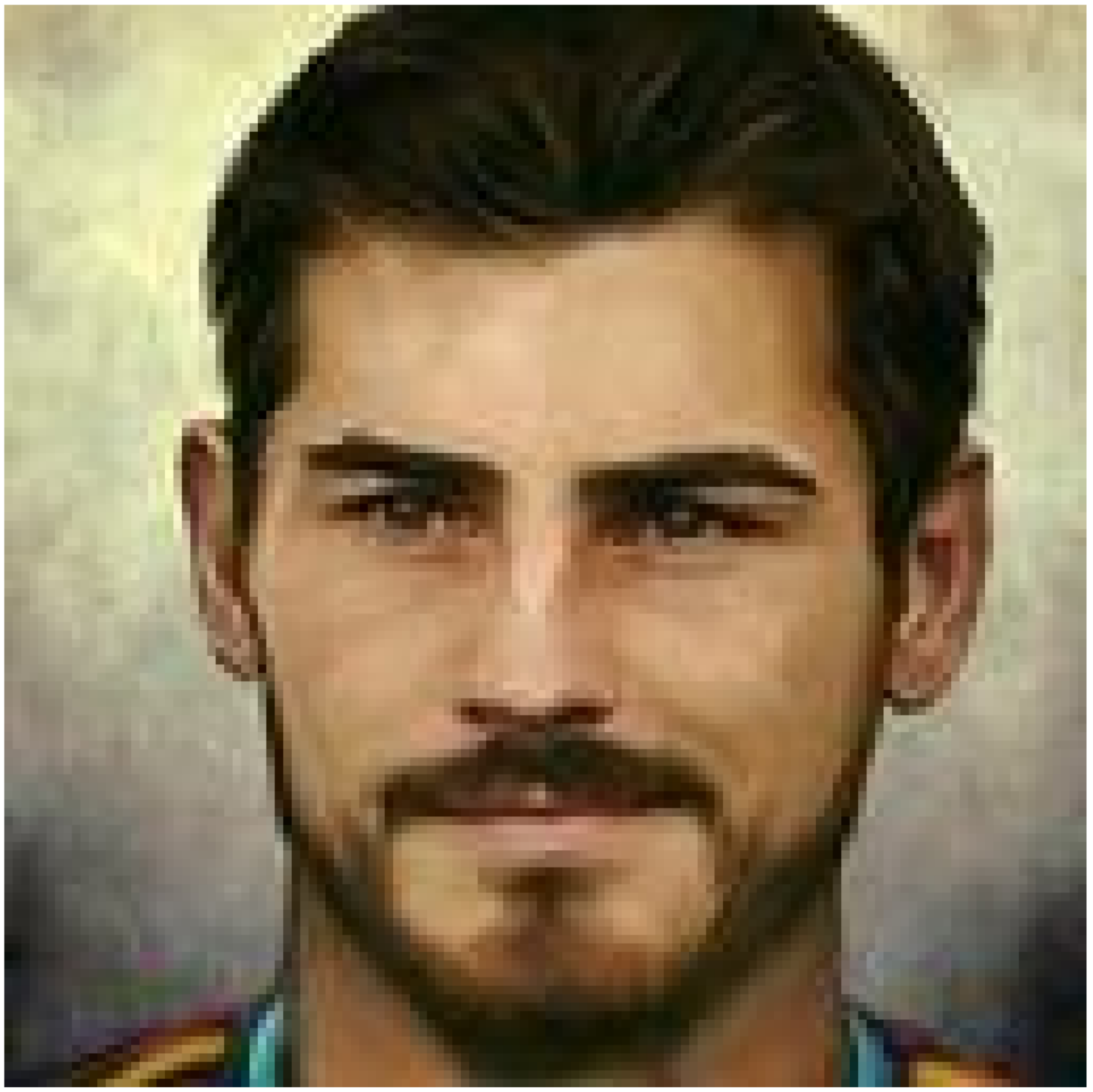}
        \label{fig:origin}
    }
    \subfigure[]{
        \includegraphics[width=0.3\linewidth]{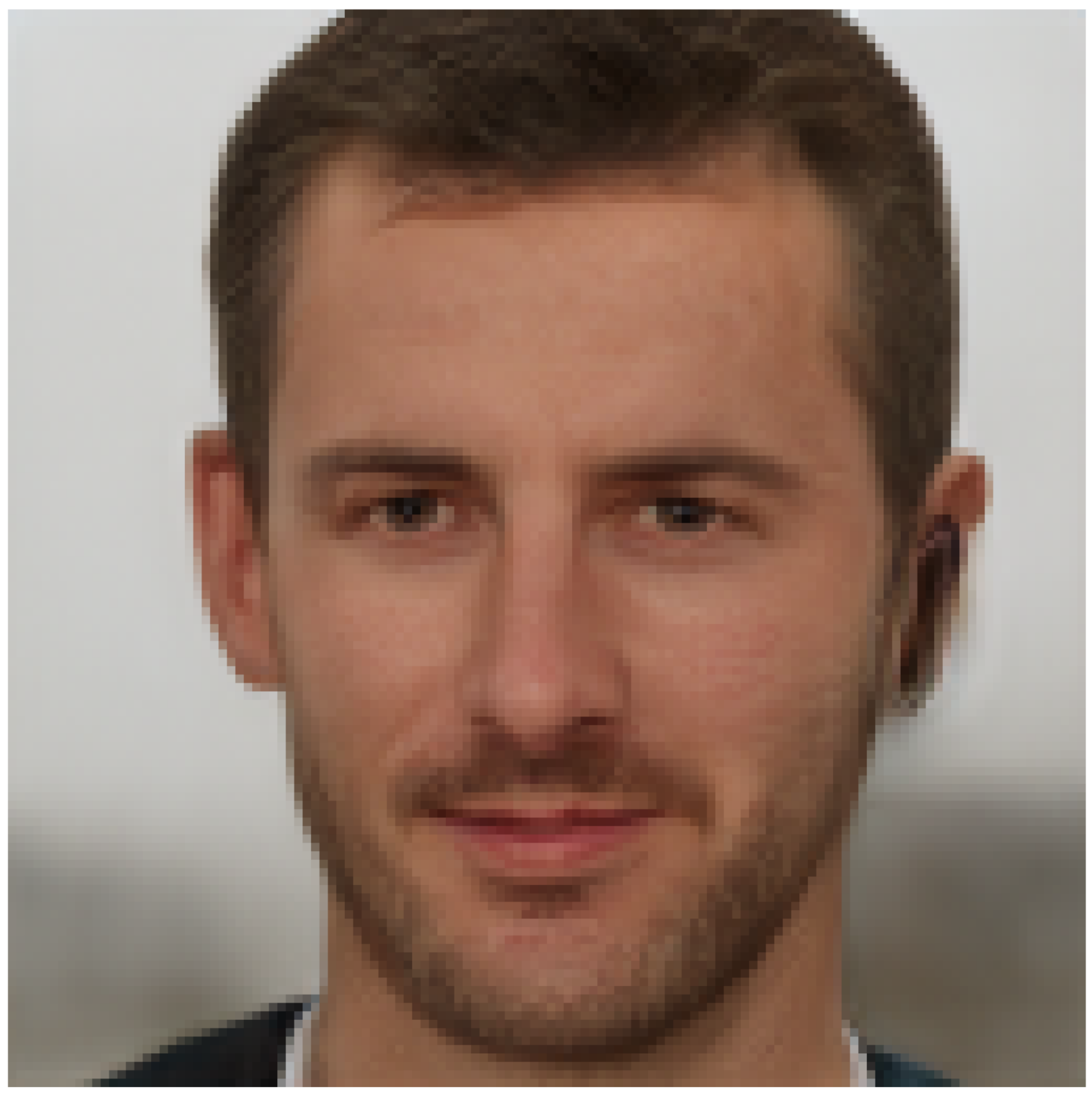}
        \label{fig:case1}
    }
    \subfigure[]{
        \includegraphics[width=0.3\linewidth]{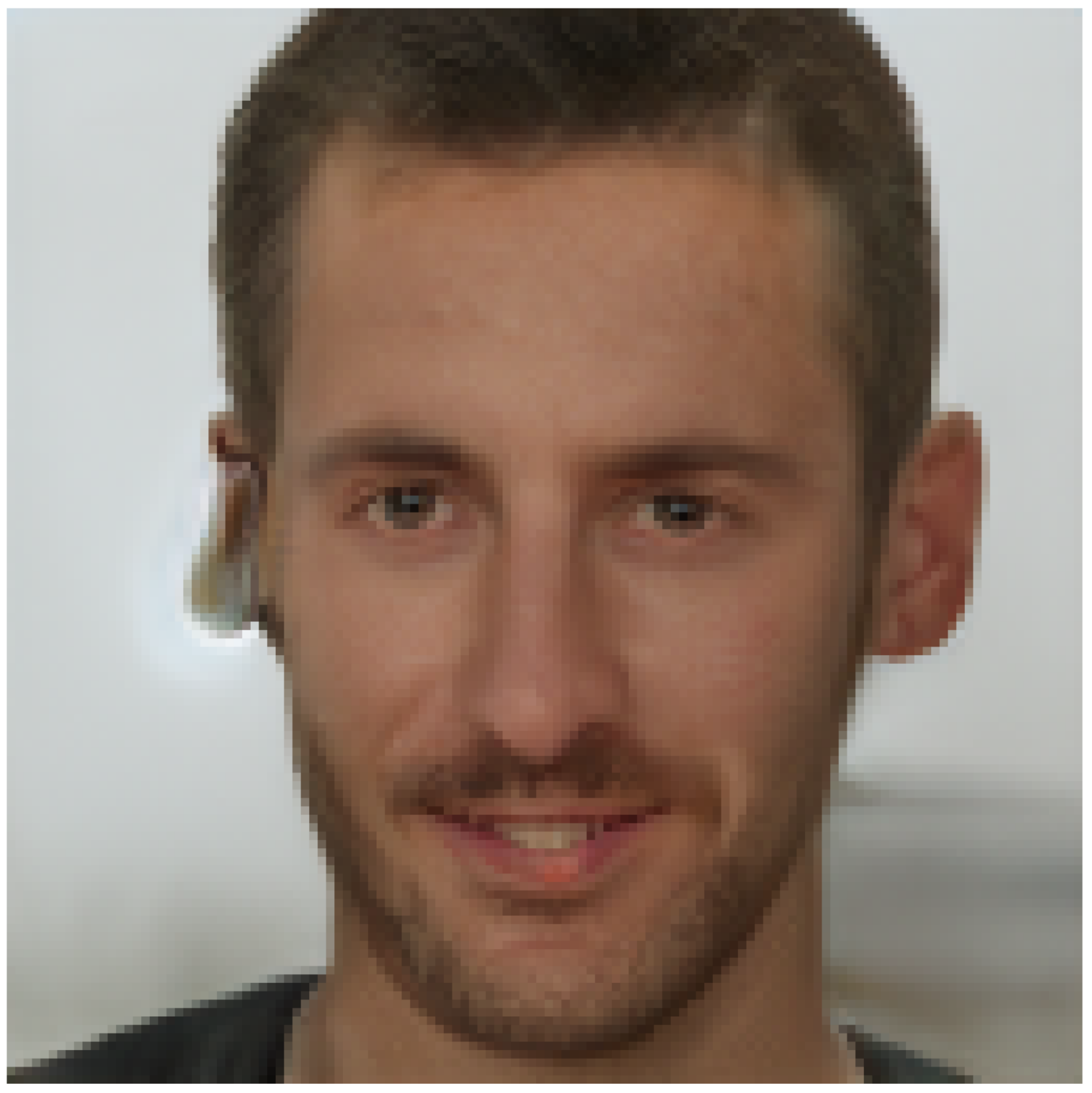}
        \label{fig:case2}
    }
   
\caption{Reasons of failing to preserve desired attributes perfectly. (a) Original image with attribute ``non-Smiling'' ; (b) Privacy protected image with the attribute``non-Smiling'' but is predicted to be ``Smiling''; (c) Privacy protected image with the attribute ``Smiling'' due to erroneous mapping to the latent space.}
\label{fig:image_observe_sameornotsame}
\end{figure}

Moreover, it is worth mentioning that the attribute preservation ratio is not $100\%$ due to the following two reasons.
We used the attribute ``Smiling'' as an example in four-attributes setting for the following discussions.
As shown in Figure \ref{fig:origin}, the original image possesses the attribute ``non-Smiling'', and this attribute is desired to be preserved after applying PI-Net.
We can visually see from Figure \ref{fig:case1} that the privacy-protected image still has the attribute ``non-Smiling''.
We find that both the original and privacy-protected images belong to the same cluster in the $\mathcal{Y}$ space.
However, the specified attribute in Figure \ref{fig:case1} is inaccurately predicted to be ``Smiling'' due to attribute prediction for attribute label assignment.

On the other hand, the original attribute ``non-Smiling'' has been changed in the resultant privacy-protected image in Figure \ref{fig:case2}, which possesses ``Smiling''.
The reason of failing to retain the desired attribute comes from the mapping error from the space $\mathcal{Y}$ to the latent space $\mathcal{X}$.

\begin{figure}[h]
    \centering
    \subfigure[2 attributes+MSE]{
        \includegraphics[width=0.3\linewidth]{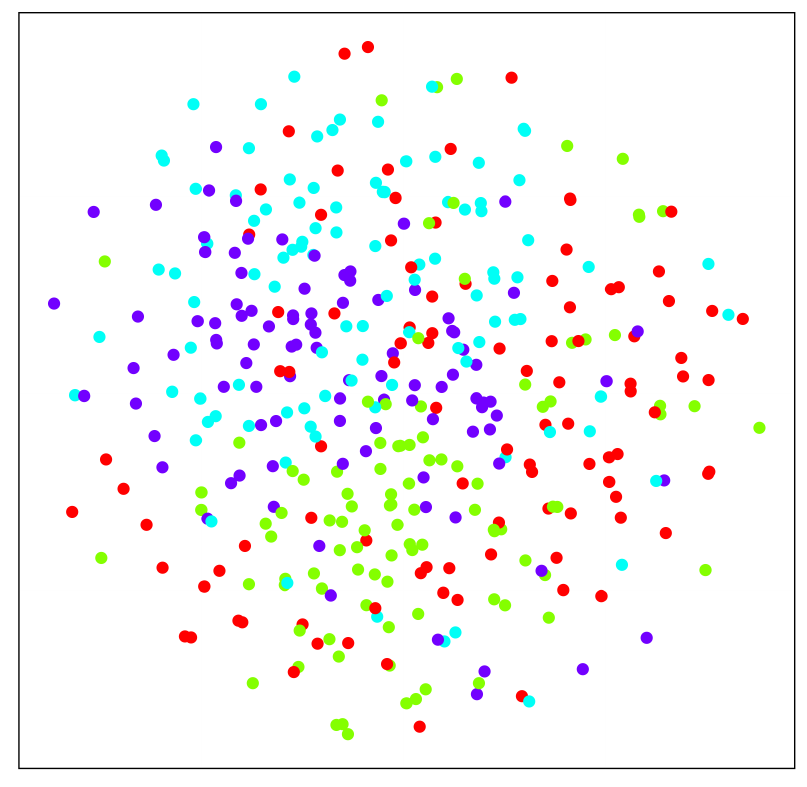}
        \label{fig:tsne_wo_2}
    }
    \subfigure[3 attributes+MSE]{
        \includegraphics[width=0.3\linewidth]{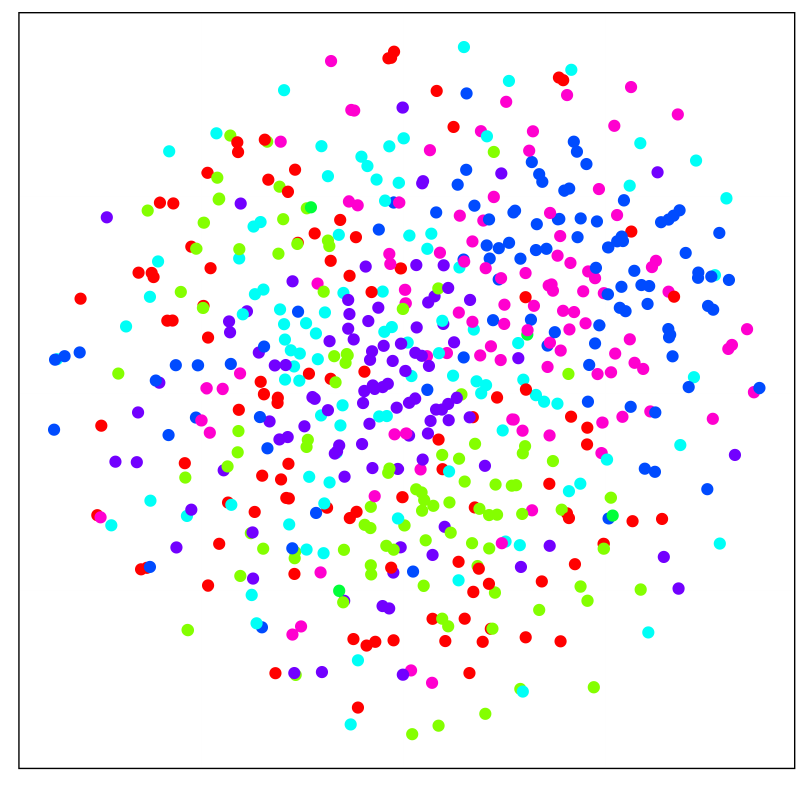}
        \label{fig:tsne_wo_3}
    }
    \subfigure[4 attributes+MSE]{
        \includegraphics[width=0.3\linewidth]{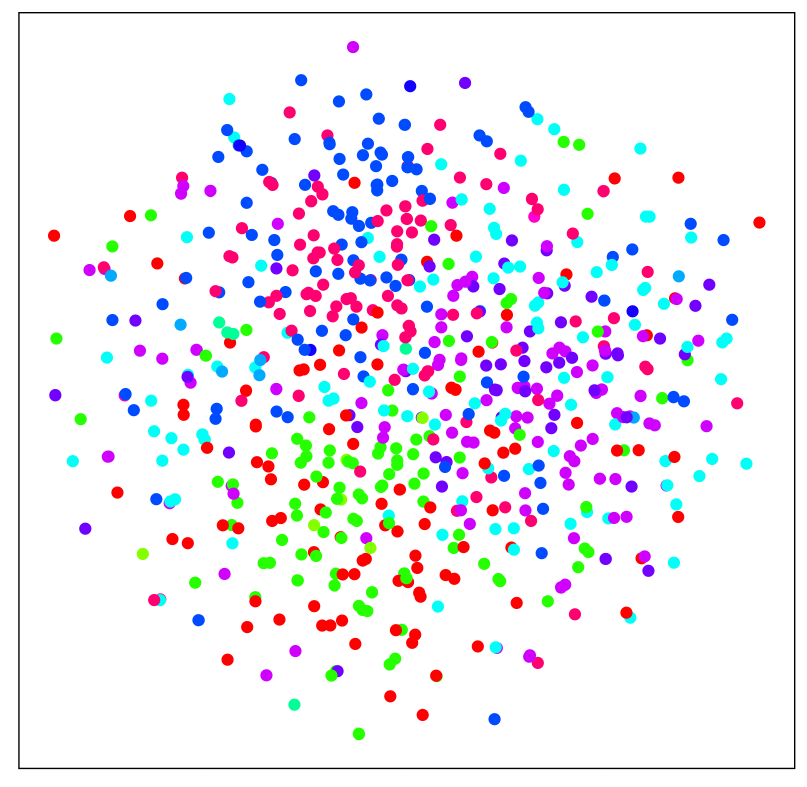}
        \label{fig:tsne_wo_4}
    }
    \\
    \subfigure[2 attrib.+TRI+CE]{
        \includegraphics[width=0.3\linewidth]{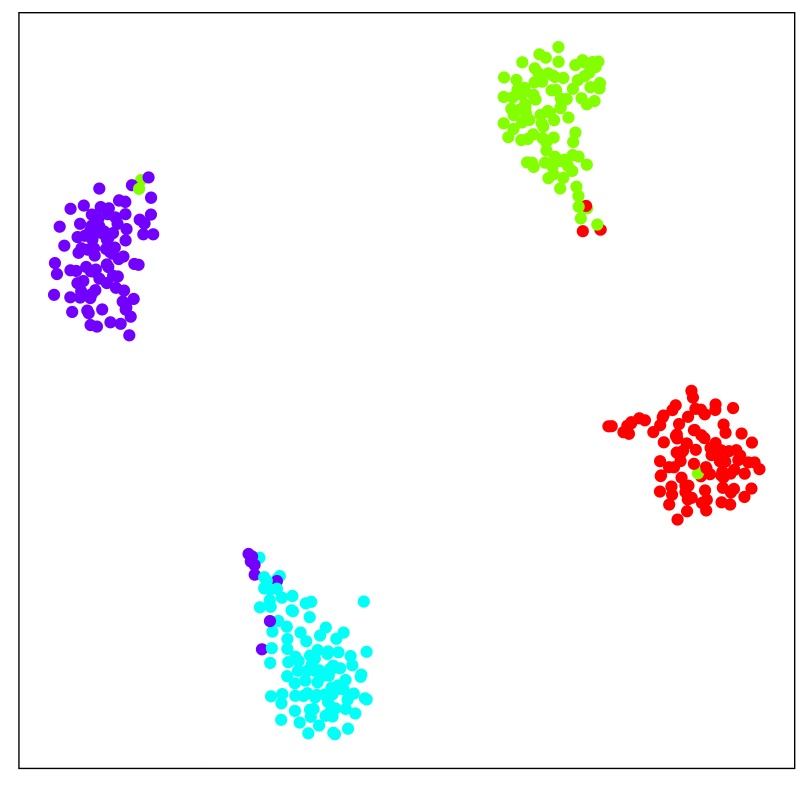}
        \label{fig:tsne_tri_ce_2}
    }
    \subfigure[3 attrib.+TRI+CE]{
        \includegraphics[width=0.3\linewidth]{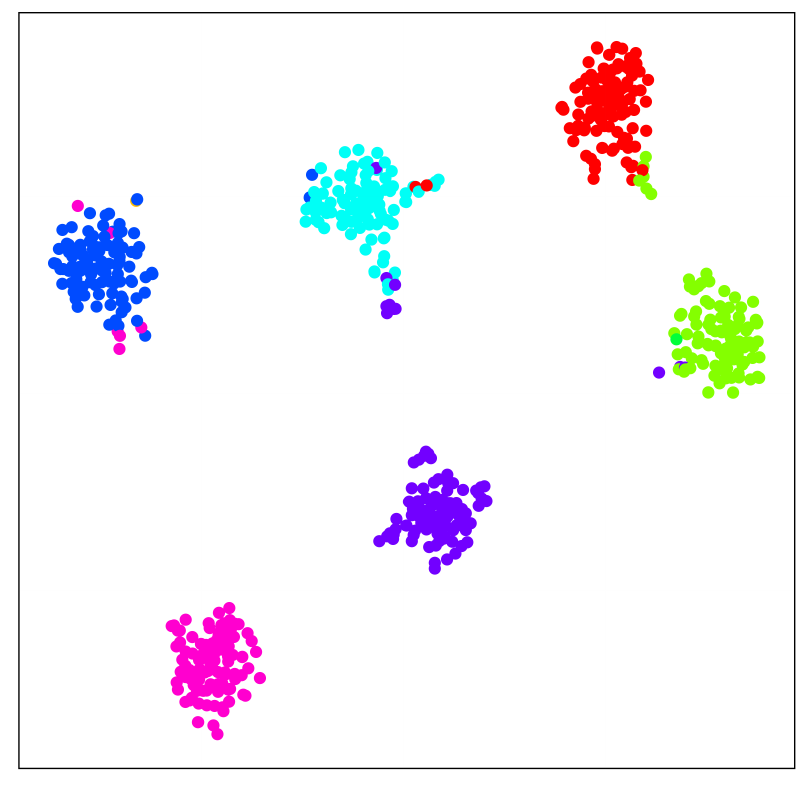}
        \label{fig:tsne_tri_ce_3}
    }
    \subfigure[4 attrib.+TRI+CE]{
        \includegraphics[width=0.3\linewidth]{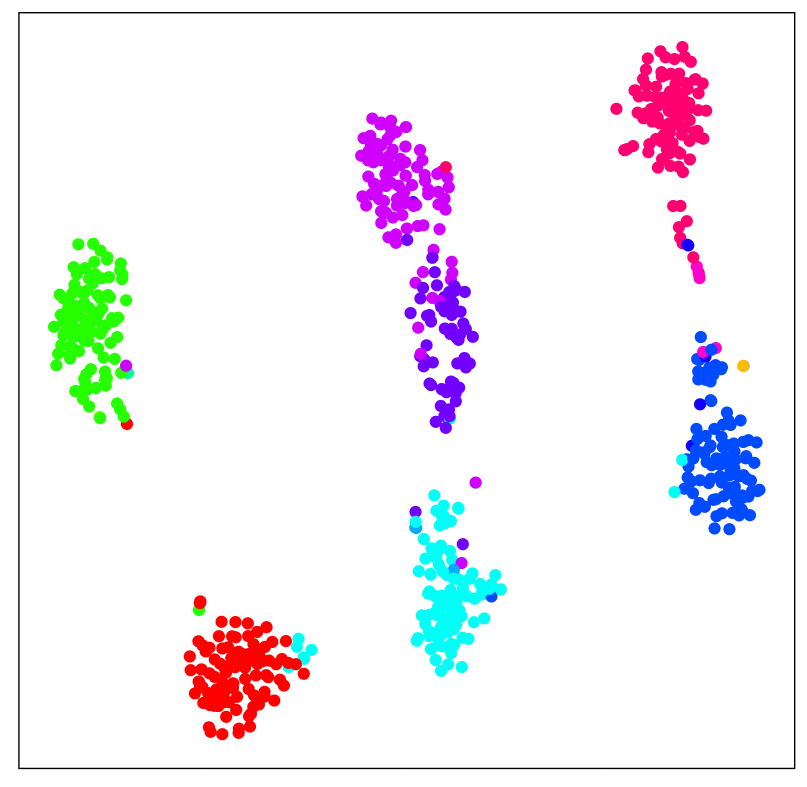}
        \label{fig:tsne_tri_ce_4}
    }
   
\caption{Visualizing the distribution of the transformed codes with $t$-SNE \cite{Maaten2008VisualizingDU}. (a)/(b)/(c) correspond to the latent code distribution with 2/3/4 attributes under the use of MSE as loss function.
(d)/(e)/(f) correspond to the latent code distribution with 2/3/4 attributes under the use of triplet and cross-entropy loss functions.}
\label{fig:tsne}
\end{figure}

\textbf{Ablation Study on Loss Functions.} Figure \ref{fig:tsne} illustrates how the triplet loss in the PI-Net enables the obfuscated image to be re-classified correctly for face attributes. 
The results in the top row were obtained in PI-Net by removing the triplet loss and keeping only the MSE loss, whereas those in the bottom row were generated using both the triplet and cross-entropy loss functions.

It can be observed from Figures \ref{fig:tsne_tri_ce_2}-\ref{fig:tsne_tri_ce_4} that because the triplet loss brings transformed codes with the same semantics close enough, the facial attributes do not deviate significantly after image obfuscation. 
Therefore, the specified facial attributes of the obfuscated images can be identical to those of the corresponding original ones.
On the contrary, for the model that only uses MSE as the loss function, the distribution of transformed codes, as shown in Figures \ref{fig:tsne_wo_2}-\ref{fig:tsne_wo_4}, becomes chaotic without forming clusters. 

Furthermore, it is worth noting that some clusters cannot be visualized because they contain too few points. This is due to
some combinations of attributes being almost non-existent.
For example, Figure \ref{fig:tsne_tri_ce_4} specifies that four binary attributes
should be preserved, but only seven clusters appear. The reason behind this is that some clusters are empty. 

\textbf{Trade-off between Privacy and Utility.} Ideally, PI-Net only obfuscates the identifiable features, while preserving the integrity of multiple face attributes.
However, due to the factors such as the noise scale introduced by PI and the imperfection of attribute clustering, PI-Net might generate the images with distortion on the attributes not selected by the user. Following \cite{maximov2020ciagan}, we also examined two important capabilities that an anonymization method should have, \emph{i.e.,} high detection rate and low identification rate.

\begin{figure}[h]
    \centering
    \includegraphics[width=0.8\linewidth]{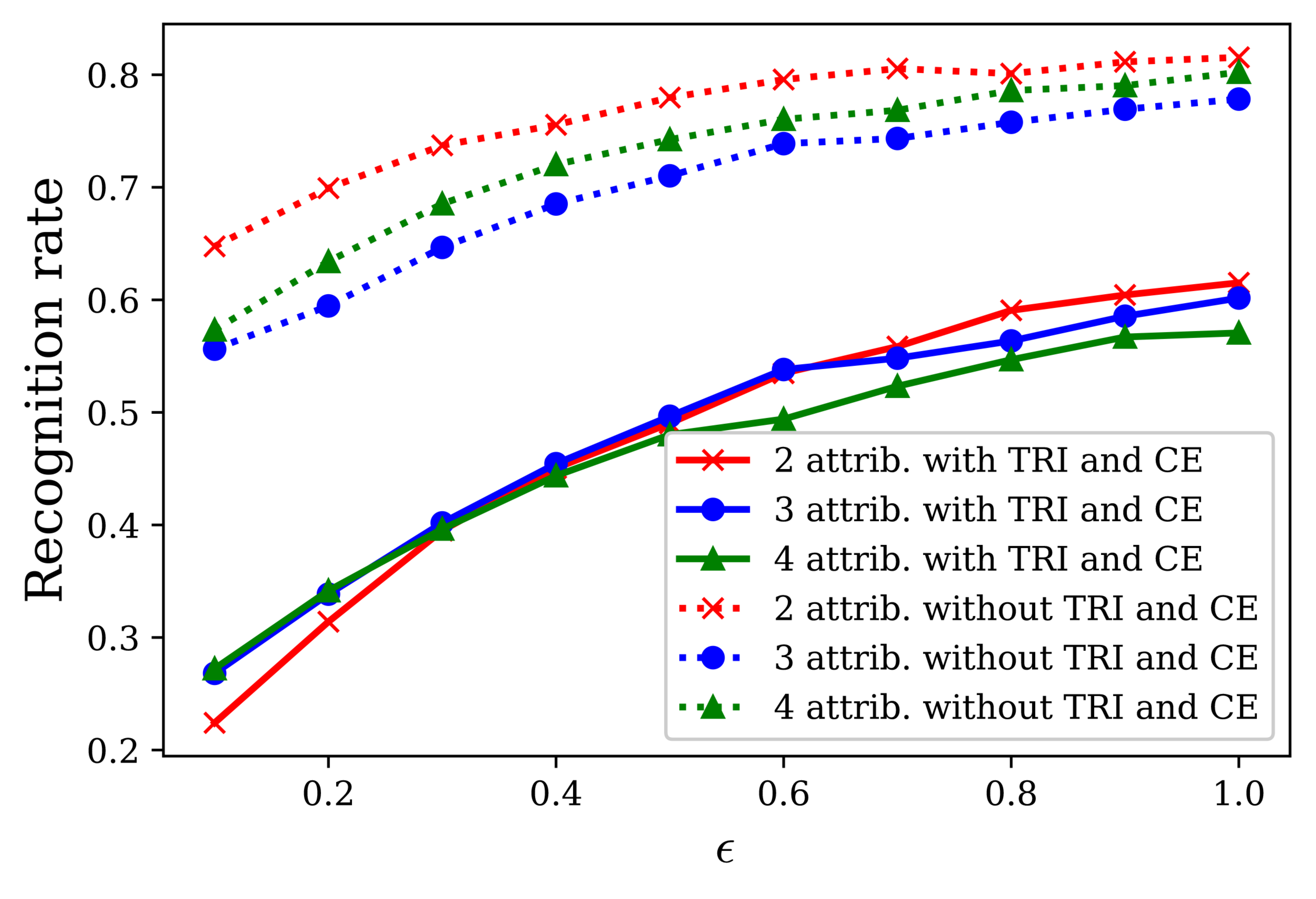}
    \caption{Re-identification ratio vs. privacy budget $\epsilon$ with/without triplet (TRI) and cross-entropy (CE) losses.}
    \label{fig:recog_rate}
\end{figure}

First, we show the trade-off between privacy and face recognition in Figure \ref{fig:recog_rate}. We used FaceNet based on Inception-Resnet backbone as our recognition model, which is released from OpenFace \cite{amos2016openface}. 
One can make the following two observations: 
(1) The more attributes that are preserved, the higher the percentage of obfuscated images that are re-identified, which means that fewer identifiable features are obfuscated. 
(2) The use of the triplet loss function and cross-entropy will increase the percentage of obfuscated images that are re-identified. This is caused by the preservation of more attributes, and therefore fewer identifiable features are available for obfuscation. To show the ability of privacy protection, the cosine similarity of Facenet embeddings could present the familiarity between origin and after protection image. Second, we examined the face detection rate of PI-Net. The results show that, unlike the conventional pixelization and blurring techniques (see Figure \ref{fig:vis} in \cite{maximov2020ciagan}), the detection rates using HOG \cite{Dalal2005hog} of PI-Net are almost $100\%$, meaning that the face structure is maintained very well. We also find that the use of triplet and cross-entropy loss functions only sacrifices negligible detections. When compared with PixDP 
\cite{Fan2019DifferentialPF} and SVD-Priv \cite{Fan2019PracticalIO}, Table \ref{tab:detect-tbl} shows that these two methods fail to preserve face image quality effectively, their face detection rates are remarkably lower than ours under different settings of privacy budgets.

\begin{table}[h]
\begin{center}
\begin{tabular}{ccccc}
                                  & 
          \textbf{$\epsilon=0.1$} &
          \textbf{$\epsilon=0.3$} &
          \textbf{$\epsilon=0.5$} &
          \textbf{$\epsilon=1.0$} \\
\toprule
PixDP     &     0.00&     0.00&     0.00&     0.00\\
\midrule
SVD-Priv  &0.11&0.18&0.25& 0.39\\
\midrule
PI-Net      &0.99&1.00&   1.00&    1.00\\
\bottomrule
\vspace{2mm}
\end{tabular}
\caption{Face detection rate (higher rate implies better preservation of a face in obfuscated image) vs. privacy budget (lower $\epsilon$ implies better privacy)
\vspace{-10mm}
}
\label{tab:detect-tbl}
\end{center}
\end{table}


\section{Conclusion}
In this paper, we have proposed Perceptual Indistinguishability-Net (PI-Net) to study how differential privacy (DP) can be employed for facial image obfuscation while maintaining certain utility.
In PI-Net, perceptual indistinguishability is presented to define the adjacency between images in the latent space for injecting noises to achieve obfuscation, and facial semantic attributes are manipulated to generate realistic looking faces.
Experimental results demonstrate that our method can satisfy the trade-off between privacy and utility.
To our knowledge, we are first to introduce perceptual similarity to define image indistinguishability in the context of DP.

\vspace{-1mm}
{\paragraph{Acknowledgements} This work was supported by Ministry of Science and Technology, Taiwan, ROC, under Grants MOST 109-2221-E-001–023, 107-2221-E-001-015-MY2, and 110-2636-E-009-018.}
{\small
\bibliographystyle{ieee_fullname}
\bibliography{biography}
}


\end{document}


\title{Appendix of Perceptual Indistinguishability-Net (PI-Net): \\
Facial Image Obfuscation with Manipulable Semantics}

\author{Jia-Wei Chen$^{1,3}$ \hspace{3mm} Li-Ju Chen$^{3}$ \hspace{3mm} Chia-Mu Yu$^{2}$ \hspace{3mm} Chun-Shien Lu$^{1,3}$\\
$^{1}$Institute of Information Science, Academia Sinica \hspace{3mm} $^{2}$National Yang Ming Chiao Tung University\\
$^{3}$Research Center for Information Technology Innovation, Academia Sinica\\

{\tt\small \{jiawei, lijuchen\}@citi.sinica.edu.tw \hspace{3mm} chiamuyu@nycu.edu.tw \hspace{3mm} lcs@iis.sinica.edu.tw}
}

\onecolumn
\maketitle
\appendix

As a number of notations are used throughout the paper, their definitions are summarized in the table below. 

\begin{table*}[ht]
\centering
\begin{adjustbox}{max width=0.95\textwidth}
\begin{tabular}{|l|l|} 
\hline
Notation                                      & Definition                                                                                                                         \\ 
\hline
$m$                                           & Number of facial attributes                                                                                                        \\ 
\hline
$J$                                           & Number of clusters                                                                                                                 \\ 
\hline
$N$                                           & Number of samples                                                                                                                  \\ 
\hline
$\mathcal{S} \subset \mathbb{R}^{m}$          & The semantic space that formed by $m$ attributes                                                                                     \\ 
\hline
$\mathcal{X} \subset \mathbb{R}^{d}$          & The latent space that learned in GANs                                                                                              \\ 
\hline
$\mathcal{Y} \subset \mathbb{R}^{k}$          & The space of transformed code                                                                                                      \\ 
\hline
$\mathcal{I}$                                 & The image space                                                                                                                    \\ 
\hline
$\mathcal{Z}$                                 & The set of outcomes from the probability function                                                                                              \\ 
\hline
$\mathcal{D}$                                 & The training dataset                                                                                                               \\ 
\hline
$s \in \mathcal{S}$                           & The semantic vector with each entry indicates whether an attribute exists or not                                                       \\ 
\hline
$x \in \mathcal{X}$                           & The latent code                                                                                                                      \\ 
\hline
$x^{\prime} \in \mathcal{X}$                  & The latent code of adjacent image                                                                                                    \\ 
\hline
$x_{i}^{a} \in \mathcal{X}$                   & The anchor of the $i$-th data                                                                                                       \\ 
\hline
$x_{i}^{p} \in \mathcal{X}$                   & The positive sample of the $i$-th data                                                                                                    \\ 
\hline
$x_{i}^{n} \in \mathcal{X}$                   & The negative sample of the $i$-th data                                                                                                    \\ 
\hline
$I \in \mathcal{I}$                           & The input image                                                                                                                    \\ 
\hline
$d_{\mathcal{X}}$                             & The distance metric for latent space $\mathcal{X}$                                                                                 \\ 
\hline
$\rho(x, x^{\prime})$                         & The perceptual distance between two latent codes $x$ and $x^{\prime}$                                                              \\ 
\hline
$K$                                           & The mechanism for assigning the probability distribution to each latent code                                                              \\ 
\hline
$K(x)$                                        & The probability distribution over $x$                                                                                              \\ 
\hline
$K(x)(Z)$                                     & The probability that the obfuscated latent code belongs to the set $Z$ when the original latent code is $x$  \\ 
\hline
$H$                                           & The mechanism that satisfies $d_{\mathcal{Y}}$-privacy                                                                             \\ 
\hline
$M$                                           & The mechanism that satisfies $\Delta d_{\mathcal{X}}$-privacy                                                                      \\ 
\hline
$G$                                           & The generator in GANs                                                                                                              \\ 
\hline
$F: \mathcal{I} \rightarrow \mathcal{X}$      & The GAN inversion model that map $\mathcal{I}$ to $\mathcal{X} \in \mathbb{R}^{d}$                                                 \\ 
\hline
$f_{S}: \mathcal{I} \rightarrow \mathcal{S}$  & The semantic scoring function can evaluate each image's facial attribute components                                                \\ 
\hline
$f_A: \mathcal{X} \rightarrow \mathcal{S}$    & The trained attribute classification model, where $\mathcal{X} \in \mathbb{R}^{d}$, and $\mathcal{S} \in \mathbb{R}^{m}$           \\ 
\hline
$\mathcal{F}_{\mathcal{Z}}$                   & $\sigma$-algebra over~$\mathcal{Z}$                                                                                                \\ 
\hline
$\mathcal{P}(\mathcal{Z})$                    & The set of probability function over~$\mathcal{Z}$                                                                                          \\ 
\hline
$D_{\epsilon, k}$                             & The probability density function for sampling the noise with $\epsilon$ privacy budget in $k$-dimensional space                            \\ 
\hline
$\Delta$                                      & The sensitivity                         \\ 
\hline
$\epsilon$                                    & The privacy budget                                                                                                                 \\ 
\hline
$\mu$                                         & The marginal threshold in triplet loss                                                                                               \\ 
\hline
$\omega,\theta$                               & The model weight of decoding and encoding network                                                                                  \\ 
\hline
$\alpha, \beta$                               & The learning rate of decoding and encoding network                                                                                 \\ 
\hline
$C_{j}$                                       & The perceptual distance of $j$-th cluster                                                                                          \\
\hline
\end{tabular}
\end{adjustbox}
\label{tab:notation}
\end{table*}

\onecolumn

In Section 4, we claim that the noise injection mechanism $H$ satisfies $d_{\mathcal{Y}}$-privacy and $M$ satisfies $\epsilon$-PI, both without a proof. We particularly note that though Fan \cite{Fan2019PracticalIO} proposed the original design of $H$ but did not provide the proof. Here, we first provide a proof that $H$ satisfies $d_{\mathcal{Y}}$-privacy. Based on such a result, we provide a formal proof that $M$ satisfies $\epsilon$-PI.

\begin{lemma}
\label{lemma:dy-priv}

If $H: \mathcal{Y} \rightarrow \mathcal{P(Y)}$ samples $\boldsymbol{y}$ from a given $\boldsymbol{y}_{\mathbf{0}}$ with the following probability density function (PDF):
\begin{equation*}
    D_{\epsilon, k}\left(\boldsymbol{y}_{\mathbf{0}}\right)(\boldsymbol{y})=C_{\epsilon, k} e^{-\epsilon \cdot d_{\mathcal{Y}}\left(\boldsymbol{y}_{0}, \boldsymbol{y}\right)},
\end{equation*} 
then $H$ satisfies $d_\mathcal{Y}$-privacy, where $\mathcal{P(Y)}$ is the set of probability measures over $\mathcal{Y}$,
    $C_{\epsilon, k}=\frac{1}{2}\left(\frac{\epsilon}{\sqrt{\pi}}\right)^{k} \frac{\left(\frac{k}{2}-1\right) !}{(k-1) !},$
and $d_{\mathcal{Y}}$ is the $k$-dimensional Euclidean distance.
\end{lemma}

\begin{proof}
For unifying the symbol usage in Definition \ref{def:dx} and Definition \ref{def:pi}, we substitute $\boldsymbol{y}$ with $\boldsymbol{z} \in \mathcal{Y}$, which is the output sampled from the PDF. 
After that, we have
\begin{align*}
    D_{\epsilon, k}\left(\boldsymbol{y_0}\right)(\boldsymbol{z})
     = C_{\epsilon, k} e^{-\epsilon \cdot d_{\mathcal{Y}}\left(\boldsymbol{y_{0}}, \boldsymbol{z}\right)}.
\end{align*}
The probability of sampling the output $\boldsymbol{z}$ belonging to the set $Z$ at given $\boldsymbol{y}_{\mathbf{0}}$ can be computed as:
\begin{align*}
    H(y)(Z) = \int_{Z} D_{\epsilon, k}\left(\boldsymbol{y_0}\right)(\boldsymbol{z}) {\,d\boldsymbol{z}},
\end{align*}
where $y$ is identical to $\boldsymbol{y_0}$, $\forall Z \in \mathcal{F_Y}$, and $\mathcal{F_Y}$ is a $\sigma$-algebra over $\mathcal{Y}$.
By triangular inequality, we derive

\begin{align*}
    \int_{Z} D_{\epsilon,k}\left(\boldsymbol{y_0}\right)(\boldsymbol{z}) {\,d\boldsymbol{z}}
    &= \int_{Z} C_{\epsilon, k} e^{-\epsilon \cdot  d_{\mathcal{Y}}\left(\boldsymbol{y_{0}}, \boldsymbol{z}\right)} {\,d\boldsymbol{z}}
     \leq \int_{Z} C_{\epsilon, k} e^{-\epsilon \cdot 
    \left(
    d_{\mathcal{Y}}\left( \boldsymbol{y^{\prime}}, \boldsymbol{z} \right) - 
    d_{\mathcal{Y}}\left( \boldsymbol{y_{0}}, \boldsymbol{y^{\prime}}\right)
    \right)} {\,d\boldsymbol{z}} \\
    & = e^{\epsilon \cdot d_{\mathcal{Y}}\left(\boldsymbol{y_{0}}, \boldsymbol{y^{\prime}}\right)} 
    \int_{Z} C_{\epsilon, k} e^{-\epsilon \cdot d_{\mathcal{Y}}\left(\boldsymbol{y^{\prime}}, \boldsymbol{z}\right)} {\,d\boldsymbol{z}} \\
    & = e^{\epsilon \cdot d_{\mathcal{Y}}\left(\boldsymbol{y_{0}}, \boldsymbol{y^{\prime}}\right)} 
    \int_{Z} D_{\epsilon, k}\left(\boldsymbol{y^{\prime}}\right)(\boldsymbol{z}) {\,d\boldsymbol{z}}.
\end{align*}

Thus, we can conclude $H(y)(Z) \leq e^{\epsilon \cdot d_{\mathcal{Y}}\left(y, y^{\prime}\right)} H(y^{\prime})(Z).$ According to the definition of metric privacy (Definition \ref{def:dx}), $H$ satisfies $d_\mathcal{Y}$-privacy.

\end{proof}

\begin{theorem}
\label{thrm:pi}
If $H$ satisfies $d_{\mathcal{Y}}$-privacy, then in the case of sensitivity $0 \leq \Delta \leq 1$, $M: \mathcal{X} \rightarrow \mathcal{P(Z)}$ defined as $M(x) = (H \circ f)(x) = H(f(x))$ satisfies $\epsilon$-PI.
\end{theorem}

\begin{proof}
According to Lemma \ref{lemma:dy-priv}, $H$ has the following properties as it satisfies $d_{\mathcal{Y}}$-privacy:

    \begin{align*}
     H(y)(Z) \leq e^{\epsilon \cdot d_\mathcal{Y}(y, y^{\prime})} H(y^{\prime})(Z).
    \end{align*}

Let $y=f(x)$ and $y^{\prime}=f(x^{\prime})$ , we have 
\begin{align*}
    \ln \left| \frac{H(f(x))(Z)}{H(f(x^{\prime}))(Z)}\right| \leq \epsilon \cdot d_\mathcal{Y}(y, y^{\prime}).
\end{align*}

Next, by substituting $H(f(x))$ with $M(x)$ and according to the definition of $\Delta$-sensitivity (Definition \ref{def:sensitive}), we have
\begin{align*}
    \ln \left| \frac{M(x)(Z)}{M(x^{\prime})(Z)}\right| \leq \epsilon \cdot \Delta d_\mathcal{X}(x, x^{\prime}).
\end{align*}

Therefore, in the case of $0 \leq \Delta \leq 1$, we can derive:
\begin{align*}
    \ln \left| \frac{M(x)(Z)}{M(x^{\prime})(Z)}\right| \leq \epsilon \cdot d_{\mathcal{X}}(x, x^{\prime}),
\end{align*}

which implies that $M$ satisfies $\epsilon$-PI according to Definition \ref{def:pi}.
\end{proof}

Intuitively, $\epsilon$-PI refers to that the closer the perceptual distance between the latent codes, the closer the probability of producing the same obfuscated output, thus making it more difficult for an adversary to distinguish between true codes. Therefore, if the perceptual distance between the two latent codes increases after the latent codes are transformed by encoding network $f$, that is, $\Delta > 1$, the probability of being distinguished by an adversary increases.

\newpage
In Theorem \ref{thrm:pi}, we proved that $M$ satisfies $\epsilon$-PI, when $H$ satisfies $d_{\mathcal{Y}}$-privacy and $0 \leq \Delta \leq 1$.
However, as discussed in Section 4, because of the need to calculate the sensitivity of each cluster $\Delta_j$, it is difficult to apply constraints in the training phase to the encoding network $f$ to achieve $0 \leq \Delta_{j} \leq 1$. 
(For brevity, the subscript $j$ for the cluster index is omitted below.)

We use the clipping function $\tilde{f}: \mathcal{X} \rightarrow  \mathcal{Y}$, which is defined as $\tilde{f}(x) = f(x^{\prime}) + \boldsymbol{g} /(\|\boldsymbol{g}\| / C)$ to bound the sensitivity of the encoding network $f$ over $x$ , where $\boldsymbol{g} = f(x) - f(x^{\prime})$ and $x$ and $x^{\prime}$ are adjacent.
That is, if $\| f(x) - f(x^{\prime})\| \geq C$, then the $f(x)$ is clipped to get $\tilde{f}(x)$ to ensure that $\| \tilde{f}(x) - f(x^{\prime})\| \leq C$, where $C = \Delta \|x - x^{\prime}\|$.
We prove that $H$ combined with the $\tilde{f}$ satisfies $\epsilon$-PI in case of configurable sensitivity being set to $0 \leq \Delta \leq 0.5$.

\begin{theorem}
\label{thrm:clip}
Given the range adjacent to latent code $x$, that is, $\|x - x^{\prime}\| \leq \beta$, if $H$ satisfies $d_{\mathcal{Y}}$-privacy and uses the $\tilde{f}$ as a clipping function, $M(x) = (H \circ \tilde{f})(x)$ satisfies $\epsilon$-PI when the configurable sensitivity is set to $0 \leq \Delta \leq 0.5$.
\end{theorem}

\begin{proof}
We first define the clipped $f(x)$ as $\tilde{f}(x)$, which implies that $\tilde{f}(x) - f(x^{\prime}) = \boldsymbol{g} /(\|\boldsymbol{g}\| / C) = \frac{f(x) - f(x^\prime) \cdot C}{\|f(x) - f(x^{\prime})\|}$.

Here $x^{\prime}$ is randomly sampled from those latent codes adjacent to $x$, to be used as an anchor point to estimate the local sensitivity around $x$. 
We substitute $x^{\prime}$ with $x^{a}$ for the anchor point to avoid subsequent semantic conflicts. 
Thus we have 
\begin{align*}
    \tilde{f}(x) = f(x^{a}) + \frac{f(x) - f(x^{a}) \cdot C_{1}}{\|f(x) - f(x^{a})\|}, \quad C_{1} = \Delta \|x - x^{a}\|.   
\end{align*}

For all other latent codes $x^{\prime}$ adjacent to $x$, the clipped value of $x^{\prime}$ is given by
\begin{align*}
    \tilde{f}(x^{\prime}) = f(x^{a}) + \frac{f(x^{\prime}) - f(x^{a}) \cdot C_{2}}{\|f(x^{\prime}) - f(x^{a})\|},
    \quad C_{2} = \Delta \|x^{\prime} - x^{a}\|.   
\end{align*}

Let $\boldsymbol{\hat{g_1}} = \frac{f(x) - f(x^{a})}{\|f(x) - f(x^{a})\|}$ and   $\boldsymbol{\hat{g_2}} = \frac{f(x^{\prime}) - f(x^{a})}{\|f(x^{\prime}) - f(x^{a})\|}$ be normalized unit vectors. Based on triangle inequality, we derive 
\begin{align*}
    \| \tilde{f}(x) - \tilde{f}(x^{\prime}) \|  
    = \| C_1 \cdot  \boldsymbol{\hat{g_1}} - C_2 \cdot \boldsymbol{\hat{g_2}} \| 
    \leq C_1 \|\boldsymbol{\hat{g_1}}\| + C_2 \|\boldsymbol{\hat{g_2}}\|
    = C_1 + C_2 \leq 2 \Delta \beta.
\end{align*}

Let $c = \tilde{f}(x)$ and $c^{\prime} = \tilde{f}(x^{\prime})$, 
we can quickly follow the proofs of Lemma \ref{lemma:dy-priv} and Theorem \ref{thrm:pi} to derive  
\begin{align*}
\int_{Z} C_{\epsilon, k} e^{-\epsilon \cdot d_{\mathcal{Y}}\left(\boldsymbol{c_{0}}, \boldsymbol{z}\right)} {\,d\boldsymbol{z}} 
& \leq  \int_{Z} C_{\epsilon, k} e^{-\epsilon \cdot 
( 
d_{\mathcal{Y}} \left(\boldsymbol{c^{\prime}}, \boldsymbol{z}\right) -
d_{\mathcal{Y}} \left(\boldsymbol{c_0}, \boldsymbol{c^{\prime}}\right)
)
}
{\,d\boldsymbol{z}} \\
& = e^{\epsilon \cdot d_{\mathcal{Y}}\left(\boldsymbol{c_{0}}, \boldsymbol{c^{\prime}}\right)} 
    \int_{Z} C_{\epsilon, k} e^{-\epsilon \cdot d_{\mathcal{Y}}\left(\boldsymbol{c^{\prime}}, \boldsymbol{z}\right)} {\,d\boldsymbol{z}},
\end{align*}

and then we have 
\begin{align*}
    H(c)(Z) \leq e^{\epsilon \cdot d_{\mathcal{Y}}\left(c, c^{\prime}\right)} H(c^{\prime})(Z).
\end{align*}

Since $c = \tilde{f}(x)$ and $c^{\prime} = \tilde{f}(x^{\prime})$, we have 
\begin{align*}
    H(\tilde{f}(x))(Z) \leq e^{\epsilon \cdot d_{\mathcal{Y}}\left(c, c^{\prime}\right)} H(\tilde{f}(x^{\prime}))(Z).
\end{align*}

After substitute $H(\tilde{f}(x))(Z)$ with $M(x)(Z)$, and based on $d_\mathcal{Y}(c, c^{\prime}) = \|\tilde{f}(x) - \tilde{f}(x^{\prime})\| \leq 2 \Delta \beta$, we derive 
\begin{align*}
    \ln \left| \frac{M(x)(Z)}{M(x^{\prime})(Z)}\right| \leq \epsilon  \cdot 2  \Delta \beta.
\end{align*}

Therefore, setting the configurable sensitivity to $0 \leq \Delta \leq 0.5$, $M$ satisfies $\epsilon$-PI.
\end{proof}